\documentclass[letterpaper]{article}

\usepackage{aaai}
\usepackage{times} 
\usepackage{helvet} 
\usepackage{courier} 

\usepackage{color}
\usepackage{amsmath}
\usepackage{amssymb}
\usepackage{amsthm}
\usepackage[ruled]{algorithm}
\usepackage{algpseudocode}
\usepackage{graphicx}
\usepackage{colortbl}
\usepackage{arydshln}

\usepackage{tikz}
\usetikzlibrary{positioning}
\usetikzlibrary{shapes.geometric}
\usetikzlibrary{shapes.symbols}
\usetikzlibrary{shadows}
\usetikzlibrary{arrows}

\title{Path Finding under Uncertainty through Probabilistic Inference}
\author {David Tolpin, Brooks Paige, Jan Willem van de Meent, Frank Wood\\University of Oxford\\ \{dtolpin,brooks,jwvdm,fwood\}@robots.ox.ac.uk}
\begin{document}

\maketitle

\begin{abstract}
  We introduce a new approach to solving path-finding
  problems under uncertainty by representing them as
  probabilistic models and applying domain-independent
  inference algorithms to the models. This approach
  separates problem representation from the inference algorithm
  and provides a framework for efficient learning
  of path-finding policies. We evaluate the new approach on the
  Canadian Traveller Problem, which we formulate as a
  probabilistic model, and show how probabilistic inference
  allows high performance stochastic policies to be
  obtained for this problem.  
\end{abstract}

\section{Introduction}

In planning under uncertainty the objective is to find the optimal
policy --- a policy that maximizes the expected reward. In most
interesting cases the optimal policy cannot be found exactly, and
approximation schemes are used to discover the policy, either
represented explicitly or as an implicit property of the planning
algorithm, through \textit{reinforcement learning}.  Approximation
schemes include value/policy iteration, Q-learning, policy gradient
methods~\cite{SB98}, as well as methods based on heuristic
search~\cite{BG01} and Monte Carlo sampling such as
MCTS~\cite{KS06,BPE+12}.

Domain-independent planning algorithms~\cite{BG01,HBG05,H06}
can be applied to different domains with little
modification, however for many applications domain-dependant
techniques are still critical in order to obtain a high
performance policy, and put the burden of
implementation on the domain expert formulating the planning
problem.

The framework of probabilistic inference~\cite{P88} proposes
solutions to  a wide range of Artificial Intelligence
problems by representing them as \textit{probabilistic
models}. Efficient domain-independent algorithms are
available for several classes of representations, in
particular for graphical models~\cite{L96}, where inference
can be performed either exactly and approximately. 
However, graphical models typically require that the full
graph of the model to be represented explicitly, and are not
powerful enough for problems where the state space is
exponential in the problem size, such as the generative
models common in planning~\cite{SKM14}.

Probabilistic programs~\cite{GMR+08,MSP14,WVM14} can
represent arbitrary probabilistic models, efficient
approximate inference algorithms have recently
emerged~\cite{WSG11,WVM14,PWD+14}. In addition to expressive
power, probabilistic programming separates modeling and
inference, allowing the problem to be specified in a simple
language which does not assume any particular inference
technique.

In this paper, we show a connection between probabilistic
inference and path finding, which allows  many path-finding
problems to be cast as inference problems using probabilistic
programs. Based on this connection, we provide a generic scheme
for expressing a path-finding problem as a probabilistic program
that infers the path-finding policy. We illustrate this generic
scheme by its application to the Canadian Traveller
Problem~\cite{PY91,BS91,NK08}. In the empirical evaluation, we
show that high performance stochastic policies can be obtained
using domain-independent inference techniques.  In the
concluding section, we discuss other possible areas of
application of probabilistic programming in planning, as well as
possible difficulties.

\section{Preliminaries}

\subsection{Probabilistic Programming}

Probabilistic programs are regular programs
extended by two constructs~\cite{GHNR14}:
\begin{itemize}
    \item The ability to draw random values from probability
        distributions.
    \item The ability to condition values computed in the
        programs on probability distributions.
\end{itemize}
A probabilistic program implicitly defines a probability
distribution over the program's output.  Formally, we define
a probabilistic program as a stateful deterministic
computation $\mathcal{P}$ with the following properties:

\begin{itemize}
\item Initially, $\mathcal{P}$ expects no arguments.
\item On every call, $\mathcal{P}$ returns either a distribution $F$, a
    distribution and a value $(G, y)$, a value $z$, or $\bot$.
\item Upon returning $F$, $\mathcal{P}$ expects a value $x$ drawn from $F$
as the argument to continue.
\item Upon returning $(G, y)$ or $z$, $\mathcal{P}$ is invoked again
    without arguments.
\item Upon returning $\bot$, $\mathcal{P}$ terminates.
\end{itemize}

A program is run by calling $\mathcal{P}$ repeatedly until
termination.  Every run of the program implicitly produces a
sequence of pairs $(F_i, x_i)$ of distributions and drawn from
them values of latent random variables. We call this sequence a
\textit{trace} and denote it by $\pmb{x}$.  A trace induces a
sequence of pairs $(G_j, y_j)$ of distributions and values of
observed random variables.  We call this sequence an
\textit{image} and denote it by $\pmb{y}$. We call a sequence of
values $z_k$ an \textit{output} of the program and denote it by
$\pmb{z}$.  Program output is deterministic given the trace.

By definition, the probability of a trace is proportional to the
product of the probability of all random choices $\pmb{x}$ and
the likelihood of all observations $\pmb{y}$:
\begin{equation}
    p_{\mathcal{P}}(\pmb{x|y}) \propto \prod_{i=1}^{\left|\pmb{x}\right|}
    p_{F_i}(x_i) \prod_{j=1}^{\left|\pmb{y}\right|}p_{G_j}(y_{j})
  \label{eqn:p-trace}
\end{equation}
The objective of inference in probabilistic program $\mathcal{P}$
is to discover the distribution of $\pmb{z}$.

Several implementations of general probabilistic programming
languages are available~\cite{GMR+08,MSP14,WVM14}. Inference is
usually performed using Monte Carlo sampling algorithms for
probabilistic programs~\cite{WSG11,WVM14,PWD+14}. While some
algorithms are better suited for certain inference types, most
can be used with any valid probabilistic program.

\subsection{Canadian Traveller Problem}

Canadian Traveller Problem (CTP) was introduced in \cite{PY91}
as a problem of finding the shortest travel distance in a graph
where some edges may be blocked.  There are several variants of
CTP~\cite{BS91,NK08,BFS09}; here we consider the
\textit{stochastic} version. In the stochastic CTP we are given
\begin{itemize}
    \item Undirected weighted graph $G=(V,E)$.
    \item The initial and the final location nodes $s$ and $t$.
    \item Edge weights $w: E\rightarrow\mathcal{R}$.
    \item Traversability probabilities: $p_o: E\rightarrow(0,1]$.
\end{itemize}
The actual state of each edge is fixed for every problem
instance but becomes known only upon reaching one of the edge
vertices.  The goal is to find a \textit{policy} that minimizes
the expected \textit{travel distance} from $s$ to $t$. The
travel distance is the sum of weights of all traversed edges
during the travel, where traversing in each direction is counted
separately.

CTP problem is PSPACE-hard~\cite{FSB+13}, however a number of
heuristic algorithms were proposed, including high-quality
policies based on Monte Carlo methods~\cite{PKH10}. Policies
are empirically compared by averaging the distance travel over
multiple instantiations of the actual states of the edges (open
or blocked) according
to the traversal probabilities. Since the travel distance is
defined only for instance where a path between $s$ and $t$
exists, instantiations in which $t$ cannot be reached from $s$
are ignored.

A trivial travel policy is realized by traversing the problem
graph in a depth-first order until the final location is reached. The
expected travel distance of the policy is bounded from above by the
sum of weights of all edges in the graph by noticing that every
edge is traversed at most once in each direction, and at most
half of the edges are traversed on average.

\section{Duality between Path Finding and Probabilistic Inference}

We shall now show a connection between path finding and
probabilistic inference. This connection was noticed
earlier~\cite{SC90} and was used to search for the
\textit{maximum a-posteriori} probability (MAP) assignment in
graphical models using a best-first search algorithm.
Here we further extend the analogy and establish a
bilateral correspondence between inferring the
distribution defined by a probabilistic model and
learning the optimal policy in a path-finding problem.

We proceed in two steps. First, following earlier work, we
establish a connection between a MAP assignment and the
shortest path. Then, based on  this
analogy, we explain how discovering the optimal policy in a
generative model can be translated into inferring the output
distribution of a probabilistic program and vice versa.

\begin{figure}
    \centering
    \begin{tikzpicture}
	[scale=0.65,every node/.style={transform shape},node/.style={shape=circle,draw}]

  \draw (0,0) ++ (3,-1) node (x1) [node] {$(F_1,x_1)$};
  \draw (x1) ++ (4,2) node (y1) [node] {$(G_1,y_1)$};
  \draw (y1) ++ (4,-2) node (x2) [node] {$(F_2,x_2)$};

  \draw [->] (0,0) -- (x1) node [midway, sloped, above] {$-\log p_{F_1}(x_1)$};
  \draw [->] (x1) -- (y1) node [midway, sloped, above] {$-\log p_{G_1}(y_1)$};
  \draw [->] (y1) -- (x2) node [midway, sloped, above] {$-\log p_{F_2}(x_2)$};

\end{tikzpicture}
    \caption{A path in the graph of a probabilistic
    program}
    \label{fig:map-path}
\end{figure}
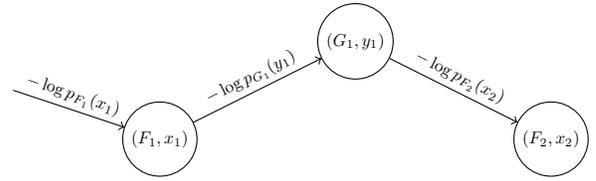

Inference on probabilistic programs computes a
representation of distribution (\ref{eqn:p-trace}).
An equivalent form of (\ref{eqn:p-trace}) is obtained by
taking logarithm of both sides:
\begin{equation}
    \log p_{\mathcal{P}}(\pmb{x}) = \sum_{i=1}^{\left|\pmb{x}\right|}
    \log p_{F_i}(x_i) +
    \sum_{j=1}^{\left|\pmb{y}\right|}\log p_{G_j}(y_{j}) + C
  \label{eqn:log-p-trace}
\end{equation}
where C is a constant that does not depend on $\pmb{x}$.
To find the MAP assignment $\pmb{x}_{MAP}$, one must maximize
$\log p(\pmb{x})$. One can view $\pmb{x}$ as a specification of
a path in a graph where each node corresponds to either $(F_i,
x_i)$ or $(G_j, y_j)$, and the costs of edges entering
$(F_i, x_i)$ or $(G_j, y_j)$  is $-\log p_{F_i}(x_i)$ o $-
\log p_{H_j} (y_j)$, correspondingly
(Figure~\ref{fig:map-path}). 
Then, finding the MAP assignment is tantamount to finding
the trace $\pmb{x}$ that produces the shortest path.
\cite{SC90,SDY07} use this correspondence in their MAP
algorithms for graphical models. 

We shall turn now to a more general case when the MAP
assignment of a part of the trace $\pmb{x}^\theta$ is
inferred.  In a probabilistic program, this is expressed by
selecting $\pmb{x}^\theta$ as the program output, $\pmb{z}
\gets \pmb{x}^\theta$. The distribution of $\pmb{z}$ is
marginalized over the rest of the trace $\pmb{x}^{\neg
\theta} = \pmb{x} \setminus \pmb{x}^\theta$, and finding the
MAP assignment for $\pmb{x}^\theta$ corresponds to finding
the mode of the output distribution:
\begin{align}
    \pmb{x}_{MAP}^\theta&=\arg \max p_{\mathcal{P}}(\pmb{z})\nonumber\\
    \propto\arg \max&
    \!\!\!\int\limits_{\pmb{x}^{\neg\theta}} \left[ \prod_{i=1}^{\left|\pmb{x^\theta}\right|} p_{F_i^{\theta}}\pmb{x}_i^{\theta} \prod_{j=1}^{\left|\pmb{y}\right|}p_{G_j}(y_{j}) \right] p_{\mathcal{P}}(\pmb{x}^{\neg \theta}) d\pmb{x}^{\neg \theta},
    \label{eqn:mode-theta}
\end{align}
where the integrand in equation (\ref{eqn:mode-theta}) depends
on $\pmb{x}^{\neg \theta}$. Just like in the case of MAP
assignment to all random variables,
equation (\ref{eqn:mode-theta}) corresponds to a path finding
problem: $\pmb{x}^\theta$ can be viewed as a policy, and
determining $\pmb{x}_{MAP}^\theta$ corresponds to learning a
policy which minimizes the expected path length
\begin{equation}
    \mathbb{E}_{\pmb{x}^{\neg \theta}}\left[-\sum_{i=1}^{\left|\pmb{x}^\theta\right|} \log p_{F_i^\theta}(x_i^\theta) 
    -\sum_{j=1}^{\left|\pmb{y}\right|}\log p_{G_j}(y_{j})\right]
    \label{eqn:log-p-trace-theta}
\end{equation}
While in principle policy learning algorithms could be used for
MAP estimation, a greater potential lies, in our opinion, in
casting planning problems as probabilistic programs and learning
the optimal policies by estimating the modes of the programs'
distributions. We suggest to adopt the Bayesian approach,
according to which prior beliefs are imposed on policy
parameters, and the optimal policy is learned through
inferring posterior beliefs by conditioning the beliefs
on observations. We explore this approach in the next
section. 

\section{Stochastic Policy Learning through Probabilistic Inference}

We have shown that in order to infer the optimum policy, a
probabilistic program for policy learning should run the agent on
the distribution of problem instance and policies, and compute
probability of each execution such that the logarithm of the
probability is equal to the negated travel cost.
\begin{algorithm}
    \begin{algorithmic}[1]
        \Require $agent$, $Instances$, $Policies$
        \State $instance$ $\gets$ \Call{Draw}{$Instances$} \label{alg:pp-learning-draw-instance}
        \State $policy$ $\gets$ \Call{Draw}{$Policies$} \label{alg:pp-learning-draw-policy}
        \State $cost$ $\gets$ \Call{Run}{$agent$, $instance$, $policy$}
        \State \Call{Observe}{1, Bernoulli$\left(e^{-cost}\right)$} \label{alg:pp-learning-observe}
        \State \Call{Print}{$policy$}
    \end{algorithmic}
    \caption{Policy learning through probabilistic inference.}
    \label{alg:pp-learning}
\end{algorithm}
The generic program shown in Algorithm~\ref{alg:pp-learning}
achieves this by randomly drawing problem instances and policies
from their distributions supplied as program arguments
(lines~\ref{alg:pp-learning-draw-instance}
and~\ref{alg:pp-learning-draw-policy}) and updating the log
probability of the sample (line~\ref{alg:pp-learning-observe})
by calling \textsc{Observe}.  \textsc{Observe} adds the log probability
of its first argument, the value, with
respect to its second argument, the distribution.
Consequently, the log probability of the output policy
\begin{align}
    \log & p_{\mathcal{P}}(policy) = \nonumber \\
         & \log p_{Policies}(policy) + \log e^{-cost(policy)} + C\nonumber \\ 
         = & \underline{- cost(policy)} + \log p_{Policies}(policy) + C
    \label{eqn:alg-pp-learning-log-p}
\end{align}
When policies are drawn from their distribution uniformly, $\log
p_{Policies}(policy)$ is the same for any policy, and does not
affect the distribution of policies specified by the probabilistic
program:
\begin{equation}
    \log p_{\mathcal{P}}(policy) = - cost(policy) + C'
    \label{eqn:alg-pp-learning-log-p-uniform}
\end{equation}
In practice, this is achieved by using a uniform distribution on
policy parameters, such as the uniform continuous or discrete
distribution for scalars, the categorical distribution with equal
choice probabilities for discrete choices, or the symmetric
Dirichlet distribution with parameter 1 for real vectors.
Alternatively, if different policies have different probabilities
with respect to the distribution $Policies$ from which the policies
are drawn, their log probabilities (taken with the opposite sign)
have the interpretation of the costs of the corresponding policies
and provide a means for specifying preferences of the model designer
with respect to different policies. In either case, the optimal
policy is approximated by estimating the mode of the program output.

When policies are drawn uniformly, the scale of the travel cost does
not affect the choice of optimal policy. However, as follows from
equation (\ref{eqn:alg-pp-learning-log-p-uniform}), the shape of the
probability density (or probability mass for discrete distributions)
depends on the cost scale --- the higher the cost, the sharper the
shape. Thus, by altering the cost scale we can affect the
performance of the inference algorithm: on one hand, the mode
estimate of a sharper function can be computed with higher accuracy,
on the other hand,  when $p_{\mathcal{P}}(policy)$ changes too fast
with its argument in the high probability region, approximate
inference algorithms converge slowly. The right scale depends on the
probabilistic program, and finding the most appropriate scale is a
parameter optimization problem.

Note that the probabilistic program for policy learning is
independent of the inference algorithm which would be used to obtain
the results. The programmer does not need to make any assumptions
about the way the mode of the output distribution is estimated, and
even whether approximate or exact inference (if appropriate) is
performed.

\section{Case Study: Canadian Traveller Problem}

We evaluated the proposed policy learning scheme on the Canadian
Traveller Problem~(Algorithm~\ref{alg:ctp}). The algorithm draws CTP
problem instances from a given graph with traversability of each
edge randomly selected according to the probabilities $p$, and
learns a stochastic policy based on depth-first search ---
the policy is specified by a vector of probabilities of selecting
each of the adjacent edges in every node.  When the policy is
realized, the selection probabilities are conditioned such that only
open unexplored edges are selected, in accordance with the base
depth-first search traversal.
\begin{algorithm}
    \begin{algorithmic}[1]
        \Require CTP$(G,s,t,w,p)$
        \State $instance$ $\gets$ \Call{Draw}{CTP$(G,w,p)$} \label{alg:ctp-draw-instance}
        \For {$v \in V$}
            \State $policy(v)$ $\gets$ \Call{Draw}{Dirichlet($\pmb{1}^{\deg(v)}$)} \label{alg:ctp-draw-policy}
        \EndFor
        \Repeat
        \State \hspace{-9pt}($reached$, $distance$) $\gets$ \Call{StDFS}{$instance$, $policy$} \label{alg:ctp-stdfs}
        \Until $reached$
        \State \Call{Observe}{1, Bernoulli$\left(e^{-distance}\right)$} \label{alg:ctp-observe}
        \State \Call{Print}{$policy$}
    \end{algorithmic}
    \caption{Learning stochastic policy for the Canadian
    traveller problem}
    \label{alg:ctp}
\end{algorithm}
$Dirichlet(\pmb{1}^{\deg(v)})$ is a uniform distribution, hence the
log probability of a trace is equal to the path cost taken with the
opposite sign.  \textsc{StDFS}~(line~\ref{alg:ctp-stdfs}) is a
flavour of depth-first search which enumerates node children in a
random order  according to the policy for the current node. An
optimal policy is expected to assign a higher probability to edges
leading to shorter paths having lower probability to become blocked.

\begin{figure}
    \centering
    \includegraphics[scale=0.4]{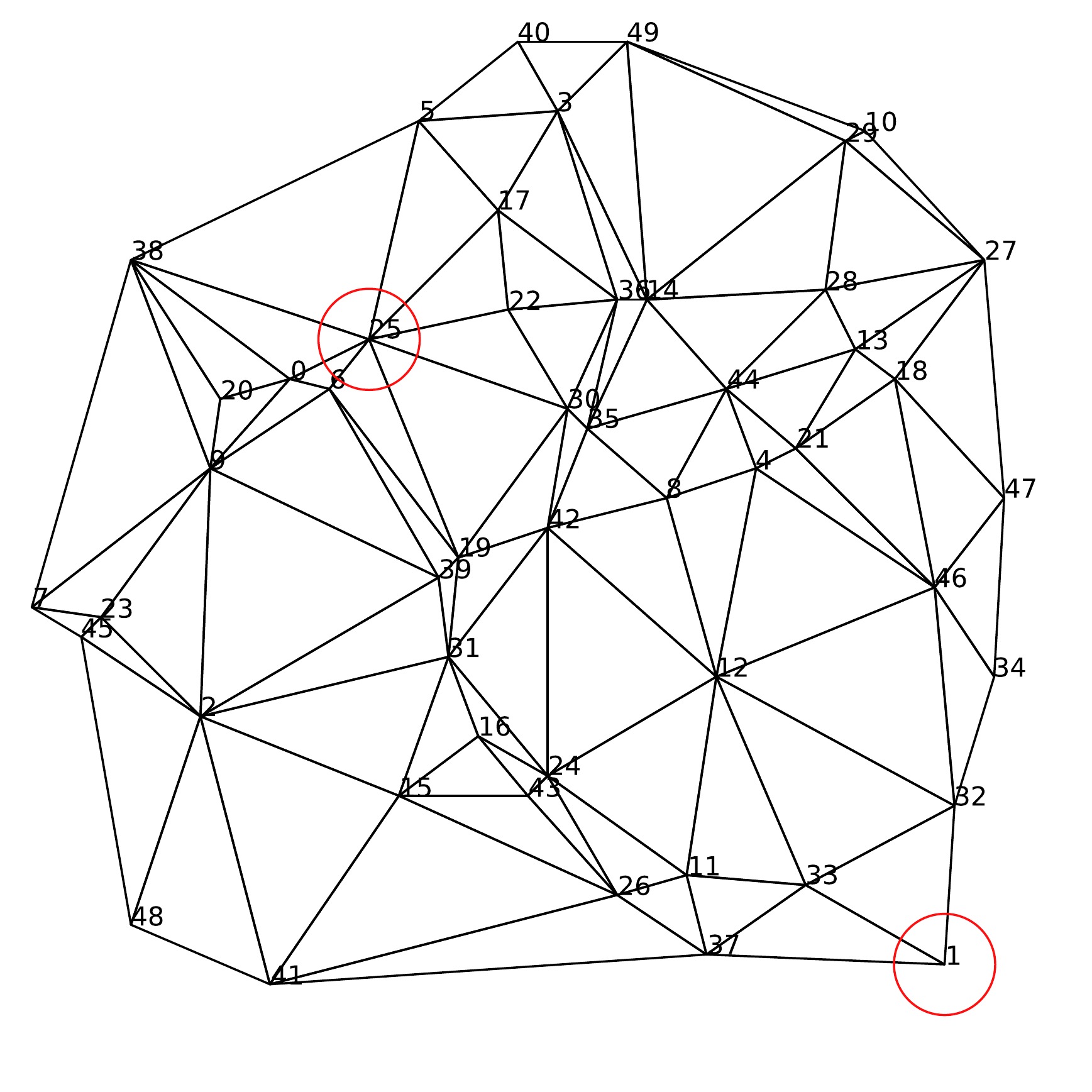}
    \caption{An instance of CTP with 50 nodes. Initial (1)
        and final (25) locations are marked by red circles;
        edge weights are Euclidean distances between edge
    vertices.}
    \label{fig:ctp-50}
\end{figure}

To assess the quality of learned policies we generated several
CTP problem specifications by triangulating a randomly drawn set
of either 50 or 20 nodes from Poisson-distributed points on a
unit square. The average DFS travel cost in fully traversable
instances was 7.9 for 50 node instances, and 5.7 for 20 node
instances. The same traversal probability in the range [0.35,
1.0] is assigned to every edge of the graph (the bond
percolation threshold for Delaunay triangulation is
$\approx$0.33~\cite{BZ09}, hence instances with $p < 0.3$ are
disconnected with high probability). A 50 node instance is shown in
Figure~\ref{fig:ctp-50}. The $s$ and $t$ nodes are marked by the
red circles, and edge weights are equal to the Euclidean
distances between the nodes.

Lightweight Metropolis-Hastings~\cite{WSG11} was used for inference.
We learned a policy for each problem specification by running the
inference algorithm for 10,000 iterations. Then, we evaluated 
policies returned at different numbers of iterations on 1,000 randomly
drawn instances to estimate the average travel cost. The average
computation time of learning and evaluation per instance 
was $\approx$80s on Intel Core i5 CPU.

\begin{figure}[t]
    \centering
    \includegraphics[scale=0.4]{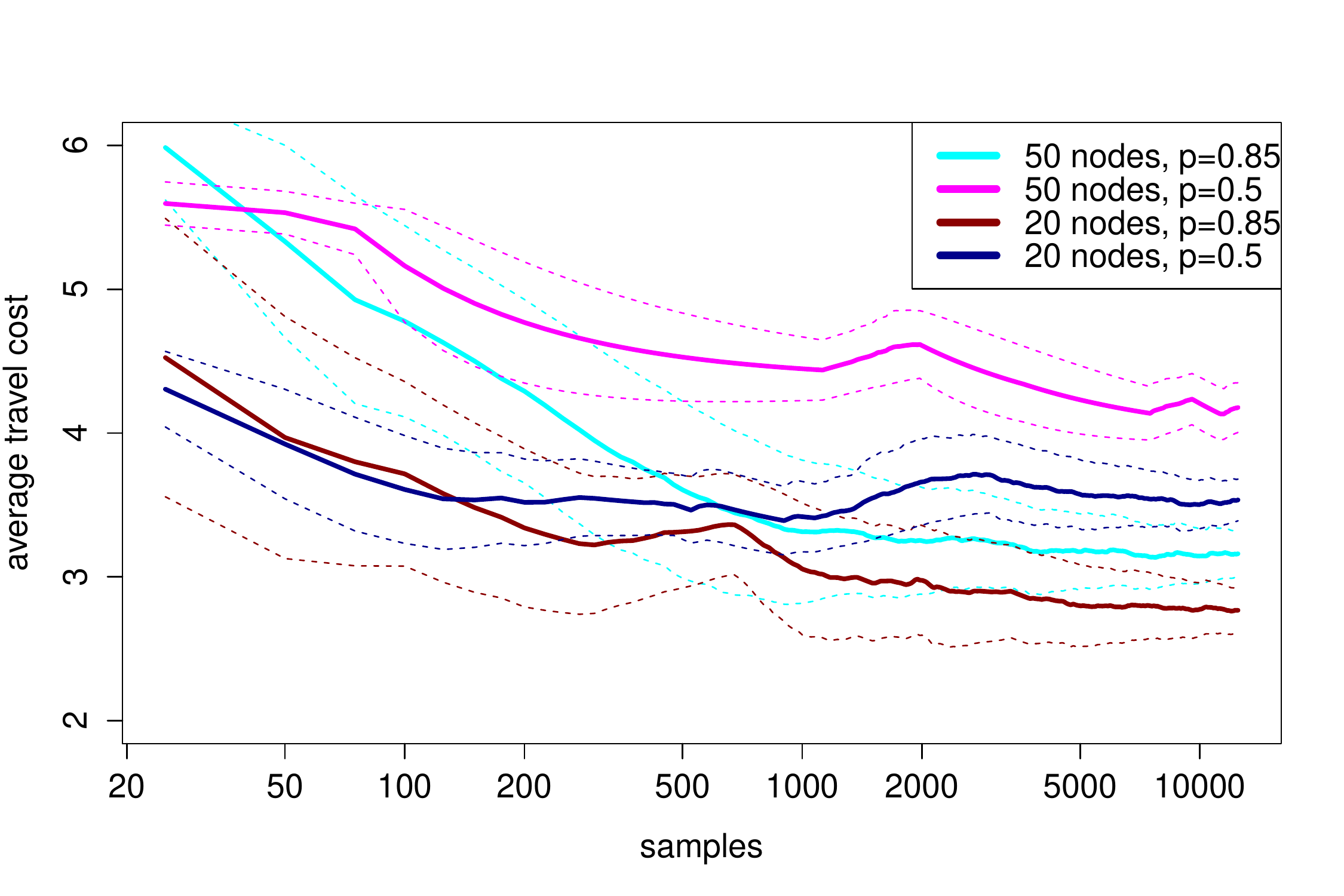}
    \caption{Average travel cost vs. number of samples for
        problems with 50 and 20 nodes and traversability
        probabilities 0.85 and 0.5. The policies mostly
        converged after $\approx$1000 samples.}
    \label{fig:atc}
\end{figure}

\begin{figure}[t!]
    \centering
    \includegraphics[scale=0.4]{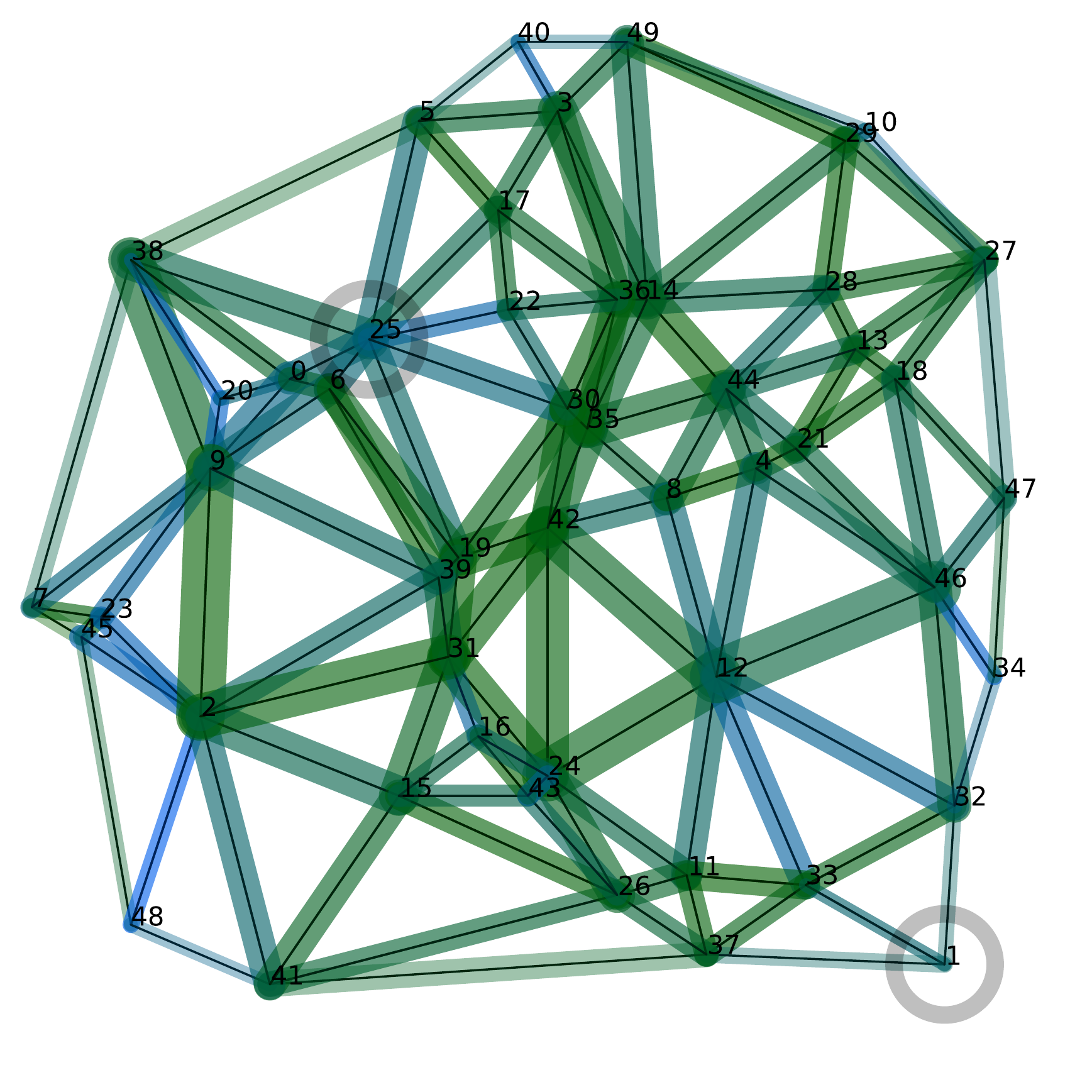}
    \caption{Visualization of policy learned for blocking
    probability 0.5 on instance in Figure~\ref{fig:ctp-50}.
    Broader edges correspond to more explored components of the
    policy.}
\label{fig:ctp-50-pol} \end{figure}

The results are shown in Figure~\ref{fig:atc}, where the solid
lines correspond to the average travel cost over the set of
problems of the corresponding size, and dashed lines to 95\%
confidence intervals. For both 50 and 20 node problems, the
policy mostly converged after $\approx$1000 iterations,
achieving 50--80\% improvement compared to the uniform
stochastic policy. While a further refinement of the policy is
possible, a different type of policy should be learned to obtain
significantly better results, for example, a deterministic
policy which takes online information into account. This,
however, would complicate the probabilistic program which we
chose to keep as simple as possible --- the actual
implementation of the program is just above 100 lines of code,
including the implementation of DFS.

A learned policy for a 50 node problem is visualized in
Figure~\ref{fig:ctp-50-pol}. Edge widths correspond to the
confidence about the policy for the edge. Edges with higher
precision (lower variance) of the policy are broader. Edge color
is blue when a traversal through the edge is much more probable in
one than in the other direction, and green when traversal in either
direction has the same probability, with shades of green and blue
reflecting how directed the edge is. As we would expect in a
converged policy, edges in the center of the graph are thicker, that
is, more explored, than at the periphery, where changes in the
policy are less likely to affect the average travel cost. Bright blue
(uni-directional) edges are mostly radial relative to the direction
from the initial position (node 1) to the goal (node 25), and many
well-explored tangential edges are green (bi-directional). This
corresponds to an intuition about the policy --- traversals through
radial edges are mostly in the direction of the goal, and through
the tangential edges in either direction to find an alternative route
when the edge leading to the goal is blocked.

\section{Discussion}

We introduced a new approach to policy learning based on casting
a policy learning task as a probabilistic program. The main
contributions of the paper are:
\begin{itemize}
    \item Discovery of bilateral correspondence between
        probabilistic inference and policy learning for path
        finding.
    \item A new approach to policy learning based on the established
        correspondence.
    \item A realization of the approach for the Canadian traveller
        problem, where improved policies were consistently learned
        by probabilistic program inference.
\end{itemize}

The proposed approach can be extended to many different planning
problems, both in well-known path-finding applications and in
other domains involving policy learning under uncertainty;
Partially observable Markov Decision Processes and generalized
Multi-armed bandit settings are just two examples. At the same
time, the exposure of probabilistic programming tools to
different domains and new applications is challenging. These
tools were initially developed with certain applications in
mind. Our limited experience shows that the probabilistic
programming paradigm scales well to new applications and larger
problems. However, as more problems are approached using the
probabilistic programming methodology, apparent weaknesses and
limitations are uncovered, and a more powerful and flexible
inference algorithm will have to be developed.

The policy learning algorithm presented here follows the offline
learning scheme --- the policy is selected before acting, and then
used unmodified until the goal is reached.  Although this is,
indeed, the easiest way to cast policy learning as probabilistic
inference, online learning can also be implemented so that when
additional computation during acting is justified by the time cost,
the policy is updated based on the information gathered online, as
in some of state-of-the-art algorithms for CTP~\cite{PKH10}.
Moreover, the time cost of updating the policy incrementally based
on the new evidence is lower than of inferring a new policy due to
the any-time nature of Bayesian updating. Online inference is a subject
of ongoing research in probabilistic programming.

By performing inference on a probabilistic program, we obtain a
representation of \textit{distribution of policies} rather than a
single policy. We then use this distribution to select a policy.
When the inference is performed approximately, which is a common
case, the expected quality of the selected policy improves with more
computation.  In the most basic setting, a fixed threshold on the
number of iterations of the inference algorithm can be imposed.  In
general, however, determining when to stop the inference and commit
to a particular policy, whether in offline or online setting, is a
rational metareasoning decision~\cite{RW91,HRT+12}. Making this decision
in an informed and systematic way is another topic for research.

\clearpage 

\bibliographystyle{aaai} \bibliography{refs}

%
%

\end{document}